\documentclass[10pt,twocolumn,letterpaper]{article}

\usepackage[pagenumbers]{cvpr} 
\usepackage{multirow}
\usepackage{graphicx}
\definecolor{cvprblue}{rgb}{0.21,0.49,0.74}
\usepackage[pagebackref,breaklinks,colorlinks,allcolors=cvprblue]{hyperref}

\def\paperID{12410} 
\def\confName{CVPR}
\def\confYear{2026}

\title{    EvRainDrop: HyperGraph-guided Completion for Effective Frame and Event Stream Aggregation    }

\author{Futian Wang$^{1}$, Fan Zhang$^{1}$, Xiao Wang$^{1}$\thanks{Corresponding Author: Xiao Wang}, Mengqi Wang$^{1}$, Dexing Huang$^{1}$, Jin Tang$^{1}$ \\ 
${^1}${School of Computer Science and Technology, Anhui University, Hefei 230601, China} \\ 
\textit{\{wft, xiaowang, tangjin\}@ahu.edu.cn}, \{e24301169, e24301148, c32314087\}@stu.ahu.edu.cn,  
}

\begin{document}
\maketitle

\begin{abstract} 
Event cameras produce asynchronous event streams that are spatially sparse yet temporally dense. Mainstream event representation learning algorithms typically use event frames, voxels, or tensors as input. Although these approaches have achieved notable progress, they struggle to address the undersampling problem caused by spatial sparsity. In this paper, we propose a novel hypergraph-guided spatio-temporal event stream completion mechanism, which connects event tokens across different times and spatial locations via hypergraphs and leverages contextual information message passing to complete these sparse events. The proposed method can flexibly incorporate RGB tokens as nodes in the hypergraph within this completion framework, enabling multi-modal hypergraph-based information completion. Subsequently, we aggregate hypergraph node information across different time steps through self-attention, enabling effective learning and fusion of multi-modal features. Extensive experiments on both single- and multi-label event classification tasks fully validated the effectiveness of our proposed framework. The source code of this paper will be released on \url{https://github.com/Event-AHU/EvRainDrop}. 
\end{abstract}

\section{Introduction} 

\begin{figure*}
\centering
\includegraphics[width=1\linewidth]{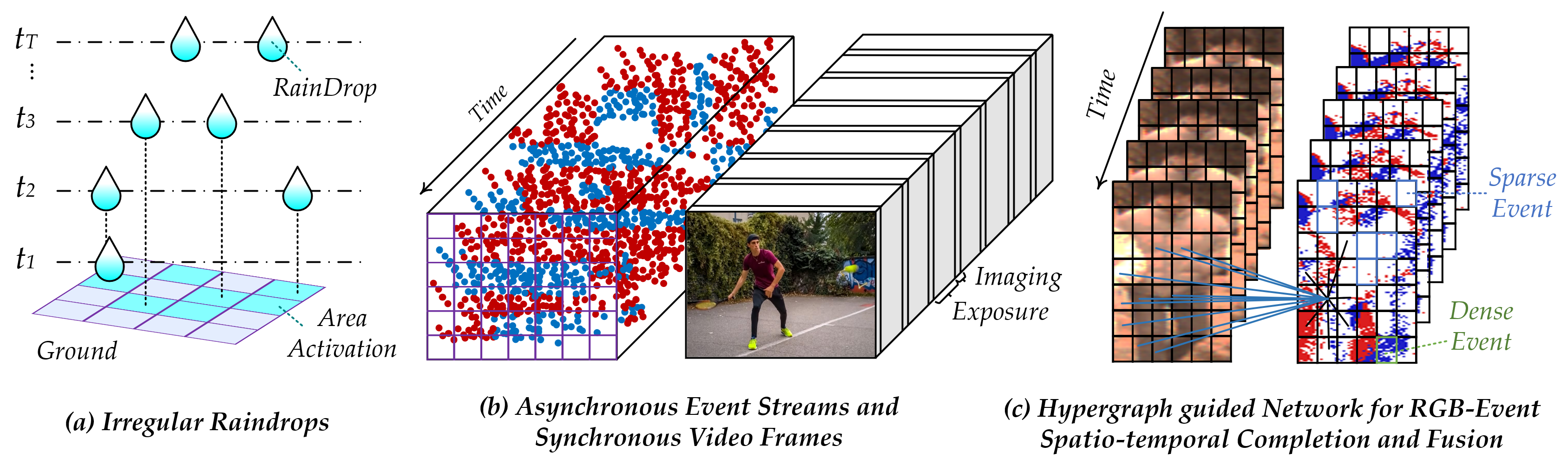}
\caption{Illustration of the irregular raindrop, irregularly sampled event streams and RGB frames, and hypergraph-guided RGB-Event spatio-temporal completion.} 
\label{fig:firstIMG}
\end{figure*}

Compared with the widely used RGB frame cameras, which capture full-frame intensity data synchronously at fixed frame rates, Event cameras, also known as Dynamic Vision Sensors (DVS),  have a significantly different biologically inspired imaging mechanism and perform better on lower energy consumption, higher dynamic range, and higher temporal resolution~\cite{rebe2021vieddo,ren2024rethinking}. As shown in Fig.~\ref{fig:firstIMG}, Event cameras capture the intensity changes in the scene, with each pixel asynchronously outputting binary information. Unlike conventional cameras, they do not produce \textit{video frames}; instead, their output consists of event streams resembling point clouds. We usually adopt a quadruple $(x, y, t, p)$ to denote an event point, where $x$ and $y$ are spatial coordinates, $t$ is the timestamp, and $p$ denotes polarity. Due to their asynchronous nature, this data is spatially sparse but temporally dense, and is free from issues such as motion blur. In addition, Event cameras also perform better in low illumination and overexposure due to their higher dynamic range.

Because of the aforementioned features and advantages, research on vision tasks centered around event cameras is gradually emerging~\cite{rebe2021vieddo}. A series of benchmark datasets are proposed for event-based pattern recognition~\cite{wang2024hardvs, wang2025sstformer, wang2024event}, object detection~\cite{wang2025object}, visual object tracking~\cite{wang2023visevent, tang2022revisiting}, and video captioning~\cite{wang2025signLTFE, wang2024eventSLT}. Specifically, Chen et al.~\cite{chen2024retain}. propose EFV++ for event stream recognition with dual-stream representation and quality-aware differentiated fusion. Wang et al.~\cite{wang2025rgb} build the EventPAR dataset and an RWKV-based framework for RGB-Event pedestrian attribute recognition. TSCFormer~\cite{wang2023unleashing} is a lightweight CNN-Transformer hybrid designed for balanced RGB-Event video recognition. The event cameras are also widely used in the low-level image processing tasks, e.g., video super-resolution~\cite{kai2025eventenhanced, kai2024evtexture}, video denoising~\cite{kar2025eventcameraguidedvisual}.

After carefully analyzing and reflecting on existing works, we find that many existing approaches~\cite{liu2021event, zubic2024state} directly stack event streams into frames. Although simple and effective, this asynchronous-to-synchronous conversion inevitably discards rich spatio-temporal information and struggles to address the inherent spatial sparsity of individual event frames. Some studies~\cite{chen2023dense, xie2024event} have attempted to use event voxels or point clouds to preserve the original spatio-temporal information better, yet the resulting performance gains remain limited. While some approaches~\cite{chen2024retain} model events as nodes in a dynamic graph with edges defined by spatio-temporal proximity, their graph construction schemes are typically sensitive and poorly scalable. This naturally raises a critical question: ``\textit{How can we design a representation for event streams that effectively preserves their rich temporal dynamics while also mitigating spatial sparsity?}"

It is well known that raindrops in nature fall asynchronously, a phenomenon that closely mirrors the inherent structure of event stream data, as illustrated in Fig.~\ref{fig:firstIMG}. 
Inspired by this phenomenon, we propose the EvRainDrop algorithm, which effectively leverages the spatiotemporal information from event streams alongside the spatial context from RGB images to mitigate the challenges caused by the spatial sparsity of event data. The key insight lies in treating asynchronous event tokens, akin to raindrops, as nodes in a hypergraph and modeling the relationships among these nodes as hyperedges. The hypergraph explicitly captures high-order dependencies among sparsely distributed events, enabling more coherent and contextually consistent feature reconstruction by modeling multi-node interactions beyond pairwise relationships. By leveraging their dense temporal information, the proposed EvRaindrop extracts reliable temporal features while simultaneously performing information completion for spatially missing tokens, guided by complementary spatial cues from the RGB modality, to yield more robust and comprehensive visual representations. After that, we introduce a self-attention mechanism to capture the global information along the time view. Extensive experiments on both single-label and multi-label classification tasks all validated the effectiveness of our proposed strategy. More detailed network architecture can be found in Fig.~\ref{fig:framework}.


To sum up, the main contributions of this paper can be summarized as the following three aspects: 

1). We propose EvRainDrop, a novel event modeling framework inspired by the asynchronous falling pattern of raindrops, which employs a hypergraph to capture high-order dependencies among sparse events for improved structural coherence.

2). We design an RGB-guided information completion strategy within the hypergraph propagation pipeline, enabling effective compensation of spatial sparsity in event data through cross-modal contextual guidance. 

3). Extensive experiments on both single-label (Human Action Recognition, HAR) and multi-label classification (Pedestrian Attribute Recognition, PAR) tasks all validated the effectiveness of our proposed framework.


\section{Related Works} 
\label{sec:relatedWorks}
In this section, we introduce the related works on the Event-based vision, Graph Neural Networks.

\subsection{Event-based Vision} 
Three primary representations have been developed for using event data in computer vision, each with trade-offs. 
Event frames convert asynchronous events into synchronous 2D images by accumulating event counts over fixed time intervals. Liu et al.~\cite{liu2021event} leveraged convolutional neural networks (CNNs) to extract spatiotemporal features from event frames and improved action recognition in dynamic scenes. Zubic et al.~\cite{zubic2024state} introduced Mamba for processing event camera signals, adapting to different sampling frequencies and achieving faster training and higher accuracy. Another representation is event point clouds, which treat individual event as 3D points $(x,y,t)$ with polarity as a feature, preserving raw spatiotemporal sparsity. Lin et al. proposed E2PNet~\cite{lin2023e2pnet}, the first learning-based event-to-point cloud registration method, which performs well under extreme lighting or fast motion. Ren et al. proposed PEPNet~\cite{ren2024simple}, which directly takes raw point clouds as input and uses a self-attention bidirectional long short-term memory module to exploit the spatiotemporal features. The third format is event voxels, which partition the spatiotemporal domain into 3D grids, balancing sparsity and structure. Chen et al. proposed E2V~\cite{chen2023dense}, which reconstructs dense 3D voxels from monocular event data and generates visually distinguishable 3D results. Xie et al. proposed EVSTr~\cite{xie2024event}, an efficient event stream spatiotemporal representation model that achieves strong performance in object classification and action recognition. 
However, these representations have inherent limitations: event frames lose fine-grained temporal details due to time discretization; event point clouds incur high computational complexity from unstructured data; and event voxels either discard local structures when the grid resolution is too coarse or introduce redundancy if non-informative voxels are retained. Our method addresses these issues by preserving rich temporal dynamics, mitigating spatial sparsity, and enabling effective spatiotemporal information completion for event streams.

\subsection{Graph Neural Network}
Graph Neural Networks (GNNs) have become a powerful tool in computer vision because they model non-Euclidean data and relational dependencies. They are effective in tasks such as capturing spatial relationships in detection~\cite{Xu2019Spatial} and modeling global context in video analysis~\cite{chen2018graphbased}. This relational reasoning also benefits multimodal learning, where GNNs fuse complementary information from diverse sources. Typical applications include integrating semantic and spatial object relationships for image captioning~\cite{yao2018exploring}, processing sparse geometric data from point clouds~\cite{chen2020hierarchical}, and learning joint representations for RGB-D salient object detection~\cite{luo2020cascade}. More recently, GNNs have been adapted to the Event-based vision domain, leveraging the sparse and asynchronous nature of event data to build efficient spatiotemporal graph representations~\cite{schaefer2022aegnn}.

Despite their success, standard GNNs mainly model pairwise relationships, where each edge connects only two nodes, which becomes a bottleneck in complex multimodal scenarios with inherently higher order interactions. Hypergraph Neural Networks(HGNNs) extend GNNs by using hyperedges that connect an arbitrary number of nodes and explicitly capture multiary higher order correlations~\cite{yang2025recent}. This capability is highly promising for multimodal learning, where early works show that hypergraphs can capture complex correlations among features from different modalities and improve recognition~\cite{mei2024MaHGNN,zhang2014Feature}. RAINDROP~\cite{zhang2021graph} builds a sensor graph for each sample, with nodes as sensors and edges as their dynamic dependencies, and propagates messages along edges with learned transmission weights to estimate the states of sensors without direct observations. Tailored for irregularly sampled multivariate time series, RAINDROP is limited by ordinary graphs in enabling refined node completion. Our EvRainDrop instead employs hypergraphs and completes nodes with missing information by exploiting the event stream and RGB frames.

\section{Methodology} \label{sec:method}

\subsection{Overview}  
Given synchronized RGB frames and event streams, both modalities are patchified, linearly projected, and augmented with positional encodings to form static (RGB) and dynamic (event) feature tokens. A hypergraph is then built over event tokens, guided by RGB spatial priors, to model high-order dependencies among sparse, asynchronous events. Two-stage hypergraph propagation is applied: 
\textit{1) intra-modal self-completion} to reconstruct missing event features via multi-node aggregation, and 
\textit{2) cross-modal enhancement} to refine them using RGB-guided hyperedges. 
Subsequently, temporal features are aggregated through self-attention along the time axis to capture long-range dynamics. Finally, fused spatio-temporal representations are fed into a classification head for pattern recognition.


\begin{figure*}
\centering
\includegraphics[width=1\linewidth]{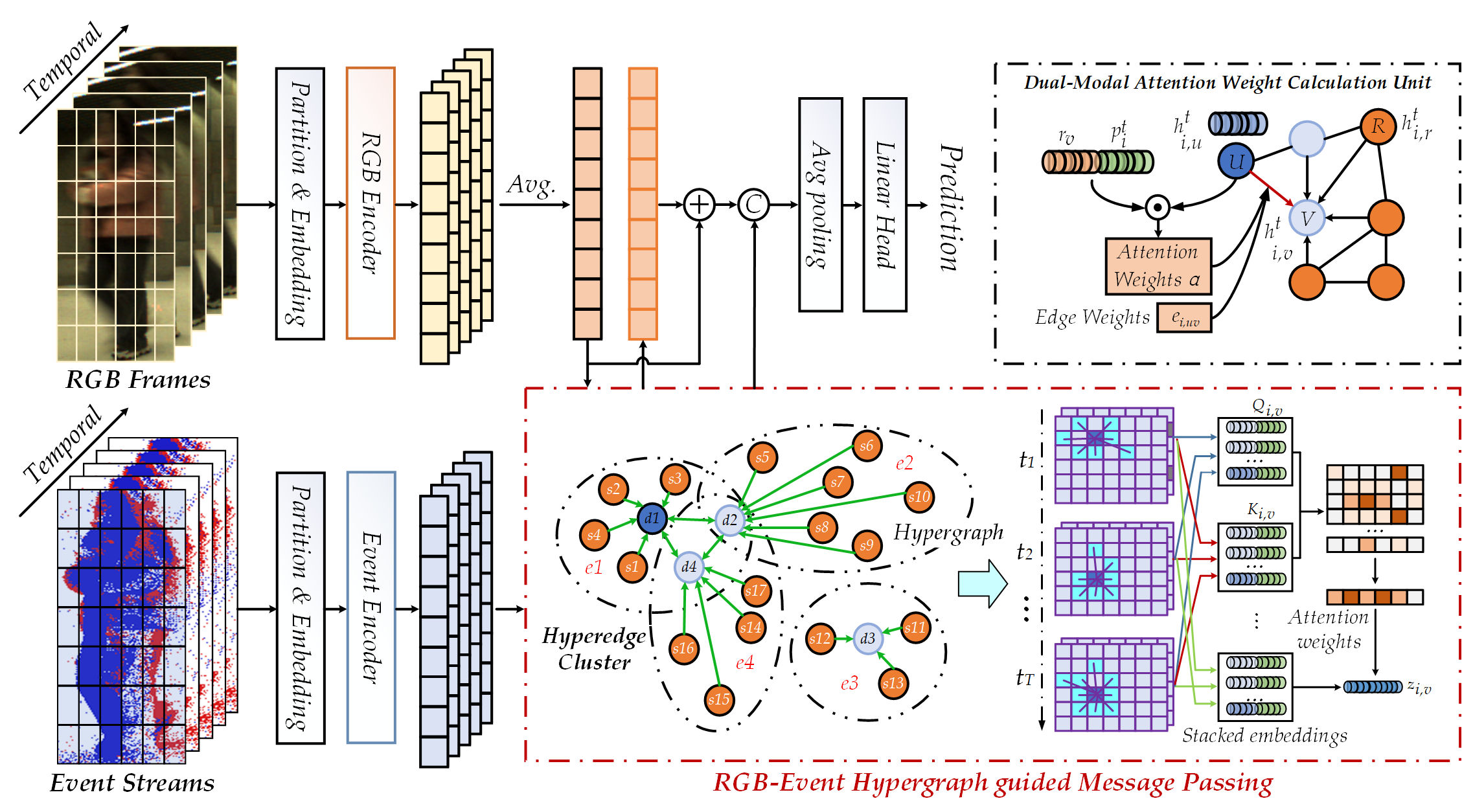}
\caption{An overview of our proposed HyperGraph-guided completion framework for effective frame and event stream aggregation, termed EvRainDrop. Specifically, we first partition and map the given RGB and event stream inputs, and then process them through a visual encoder to obtain RGB and Event tokens. Subsequently, the dual-modal tokens are fed into the hypergraph-guided message passing module, where dense event stream information and RGB spatial information are leveraged to compensate for the sparsity of event streams in the spatial domain. Afterward, we concatenate the RGB features with the enhanced event stream and pass it through a classification head to perform downstream pattern recognition, including pedestrian attribute recognition and human action recognition.}  
\label{fig:framework}
\end{figure*}

\subsection{EvRainDrop Framework} 

\subsubsection{Input Representation} 
We design an input representation process that accounts for the characteristics of both modalities while ensuring compatibility for later processing. The RGB modality uses a sequence of frames $I_r \in \mathbb{R}^{B \times T_r \times C \times H \times W}$, where $B$ is the batch size, $T_r$ is the number of RGB frames, $C$ stands for RGB channels, and $H$ and $W$ are the height and width of each frame. 
For the event stream $\mathcal{E} = \{e_1, e_2, ..., e_N\}$, we first stack them into event frames $I_e \in \mathbb{R}^{B \times T_e \times C_e \times H \times W}$, where $T_e$ is the number of event frames, $C_e$ is for polarity channels. Each frame is divided into patches, producing $N=HW/{p^2}$ tokens that are linearly projected into $D$-dimensional feature vectors. Position embeddings are added to preserve spatial information. This results in in feature sequences $I_r \in \mathbb{R}^{B \times T_r \times N \times D}$ and $I_e \in \mathbb{R}^{B \times T_e \times N \times D}$. The RGB encoder models spatial and temporal dependencies from $I_r$, while the event encoder captures the sparse and dynamic characteristics of $I_e$. To balance computational cost, we average the RGB features to get $S_f \in \mathbb{R}^{B \times S}$. For the event features, redundant information is filtered using cosine similarity, and adaptive pooling fixes the temporal dimension to $T$, yielding $D_f \in \mathbb{R}^{T \times B \times D}$.

\subsubsection{RGB-Event Hypergraph guided Message Passing}
\noindent $\bullet$ \textbf{Hypergraph Construction:} We treat each feature dimension of the RGB tokens and Event tokens as individual nodes. In what follows, we refer to Event tokens as dynamic features and RGB tokens as static features. Dynamic feature matrices $D_f \in \mathbb{R}^{T \times B \times D}$ and static feature matrices $S_f \in \mathbb{R}^{B \times S}$ are provided, where $T$ represents the number of time steps, $B$ the batch size, and $D$ and $S$ the feature dimensions. Consequently, $D$ and $S$ also denote the number of dynamic nodes and static feature nodes, respectively.

Each sample corresponds to an independent multimodal observation sequence, consisting of 
$T$ time steps of dynamic observations and static features. Formally, the $i$-th sample aggregates a sequence of temporal observations $\{D_{f,t}\}^T_{t=1}$ and a set of static feature nodes $\{S_{f}\}$. The sample thus represents one complete multimodal perception unit, where dynamic nodes provide temporally evolving event information, and static nodes supply dense spatial or contextual cues.
To model cross-modality interactions, a hypergraph is constructed for each sample individually, as shown in the left part of the RGB-Event Hypergraph guided Message Passing in Fig.~\ref{fig:framework}, with all hypergraphs kept independent within a batch.

After expanding the static feature matrix $S_f$ along the time dimension to align with the temporal features of the dynamic matrix, we map the dynamic features $D_{f}$ and the static features $S_{f}$ into a shared latent space using two separate linear projection layers. Affinity scores between dynamic nodes and static feature nodes are then computed via dot product in this shared space, which are finally normalized with the Softmax function to obtain the affiliation probabilities $A_p \in \mathbb{R}^ {S \times B \times D}$, where each element $A_p^i \in [0, 1]$.


Each dynamic node forms a hyperedge by connecting the top-$k$ static nodes with the highest probabilities. The hypergraph construction is formulated as follows:
\begin{equation}
\begin{aligned}
\mathcal{H} &= \{V, \mathcal{E}, \mathbf{H}\},
\end{aligned}
\end{equation}

\begin{equation}
\begin{aligned}
\mathbf{H} &= 
\begin{cases}
1, & \text{if } s_d^t \text{ or } d \in e, \\
0, & \text{otherwise},
\end{cases} \\[6pt]
\end{aligned}
\end{equation}
where $\mathbf{H}$ is the node–hyperedge incidence matrix, $V$ and $\mathcal{E}$ denote the vertex and hyperedge set. Here, $e$ represents a hyperedge, and 
$d$ and $s_d^t$ are dynamic and static nodes that belong to $e$, respectively. 
This hypergraph construction enables explicit interactions between dynamic and static nodes, extending beyond the pairwise relationships of conventional graphs.

\noindent $\bullet$ \textbf{Two-Stage Hypergraph Propagation:} EvRainDrop employs two cascaded propagation stages to refine both dynamic and static nodes, focusing primarily on enhancing the information-sparse dynamic nodes. These propagation operations are performed per sample, omitting hypergraph indices for simplicity.

\textbf{Hypergraph-based Message Passing.} As illustrated in the Dual-Modal Attention Weight Calculation Unit of Fig.~\ref{fig:framework}, the message passing mechanism updates each dynamic node $d$ by aggregating information from its associated nodes. As an example, we consider the static nodes $s_d^1, \dots, s_d^k$. This propagation process is observation-driven: a dynamic node becomes active and participates in message passing only when a valid observation exists at the corresponding time step. During inactive steps, the node is excluded from propagation to avoid the diffusion of empty information. In contrast, static nodes remain active throughout the process, ensuring temporal continuity and enabling robust cross-modality interactions even when dynamic observations are missing.

In this process, the dynamic node $d$ acts as the target node, while static nodes act as source nodes. The message from a source node $s_d^t$ to its target node $d$ is formulated as:
\begin{equation}
\begin{aligned}
&\:\:\:\:\:\:\:\:\:\:\:\:\: \gamma_{s_d^t} = 
\operatorname{Softmax}(e_{s_d^t}), \\
&\:\:\:\:\:\:\:\: q_d = W_Q x_d, \:\:
k_{s_d^t} = W_K x_{s_d^t}, \\
&\:\:\:\:\:\:\:\:\: \alpha_{s_d^t} = 
\operatorname{Softmax}
\!\left(
\frac{q_d^\top k_{s_d^t}}{\sqrt{d_h}}
\right), \\
& h_{s_d^t \rightarrow d} = (\gamma_{s_d^t} + \alpha_{s_d^t}) \cdot \mathrm{ReLU}(r_vx_{s_d^t}),
\end{aligned}
\label{eq:obs_path}
\end{equation}
where $x_{s_d^t}$ and $x_d$ denote the feature of $s_d^t$ and $d$. $W_Q$ and $W_K$ are learnable projection matrices used to obtain the query and key representations, and $r_v$ is a learnable linear transformation for projecting the source features into the message space. The term $\gamma_{s_d^t}$ and $\alpha_{s_d^t}$ represent the hyperedge affiliation weight and the self-attention weight, with $d_h$ is the dimensionality of each attention head.

The updated feature of the target dynamic node $d$, formulated as:
\begin{equation}
\begin{aligned}
h'_d = \displaystyle \sum_{i=1}^k h_{s_d^i \rightarrow d},
\end{aligned}
\end{equation}
is obtained by aggregating messages from all source static nodes in its hyperedge. Meanwhile, each static node also acts as a target node in the same hypergraph, receiving and aggregating messages from both the dynamic nodes and other static nodes within the hyperedge to update its representation, enabling bidirectional information flow between dynamic and static modalities.

\textbf{Stage 1: Dynamic Node Self-Completion}. This stage models the self-completion of dynamic nodes, involving only the blue nodes in Fig.~\ref{fig:framework}. A fully connected adjacency matrix $G \in \mathbb{R}^{D \times D}$ defines the initial pairwise relationships among dynamic nodes. $G$ serves as the adjacency matrix of an undirected graph $\mathcal{G}$, whose vertices are dynamic nodes and edges represent their pairwise interactions. Non-zero entries of G are converted into edge indices and weights, each forming a 2-node hyperedge connecting two dynamic nodes, one as the target and the other as the source, to align with the aforementioned hypergraph-based message passing mechanism. Dynamic features $D_f$ are updated through this mechanism, which encapsulates the relevant computations. For brevity, we denote this process as the $\operatorname{Propagate}$ function:
\begin{equation}
\begin{aligned}
D_f' = \operatorname{Propagate}(D_f, \mathcal{G})  .
\end{aligned}
\end{equation}

This step ensures that when a dynamic node generates an observation at a specific timestamp, relevant information is available at that moment, and the mechanism then propagates this information along the edges to adjacent dynamic nodes, thereby mitigating sparsity in the representation.

\textbf{Stage 2: Cross-Modality Hypergraph Enhancement}. This stage integrates static feature nodes via the hypergraph to guide and refine both modalities, encompassing the blue nodes and orange nodes in Fig.~\ref{fig:framework}. The updated dynamic features $D_f'$ and static features $S_f$ are concatenated to form the full feature matrix $C_f \in \mathbb{R}^{T \times B \times (D+S)}$. Messages are then propagated over the constructed hyperedges using the same mechanism:
\begin{equation}
\begin{aligned}
C_f' = \operatorname{Propagate}(C_f, \mathcal{H}).
\end{aligned}
\end{equation}

The updated feature matrix $C_f'$ is split into enhanced dynamic features $D_{e_f}$ and enhanced static features $S_{e_f}$. This mechanism enables bidirectional modality enhancement, where dynamic nodes benefit from static guidance while static nodes are simultaneously updated using dynamic context. Subsequently, as depicted in the right part of the RGB-Event Hypergraph guided Message Passing in Fig.~\ref{fig:framework}, we leverage the Transformer to fuse temporal information for the enhanced dynamic features. After concatenation again, the features are processed by a self-attention mechanism to obtain the representation, which is then passed through an average pooling layer and a linear classification head to generate the final prediction.




\subsection{Optimization} 
In this paper, we validate our EvRrainDrop strategy based on two classification tasks, i.e., the single-label classification (Human Activity Recognition, HAR) and the multi-label classification (Pedestrian Attribute Recognition, PAR). For the HAR problem, we build our framework based on TSCFormer~\cite{wang2023unleashing}, which adopts the cross-entropy loss function to optimize the network, i.e., 
\begin{equation}
\label{eq:cross-entropy} 
\mathcal{L}_{ce} = -\left[ y \log \hat{y} + (1 - y) \log(1 - \hat{y}) \right],
\end{equation}
where $\hat{y}$ denotes the predicted label distribution, and $y$ represents the ground truth.

For the PAR task, we adopt the weighted cross-entropy loss function based on VTB~\cite{cheng2022simple}, which can be formulated as: 
\begin{equation}
    \label{eq:wce_loss} 
    \small
    \mathcal{L}_{wce} = -\frac{1}{N} \sum_{i=1}^{N} \sum_{j=1}^{M} \omega_j \left( y_{ij} \log(p_{ij}) + (1 - y_{ij}) \log(1 - p_{ij}) \right),
\end{equation}
where $N$ is the total number of samples, $M$ is the number of attributes, and $\omega_j$ denotes the weight for the $j$-th attribute. $p_{ij}$ and $y_{ij}$ represent the predicted result and the ground truth, respectively.

\begin{table*}
\centering
\caption{Results on MARS-Attribute~\cite{zheng2016mars} and DukeMTMC-VID-Attribute~\cite{ristani2016performance} RGB-Event based PAR dataset. The best three results are highlighted in \textcolor{red}{\textbf{red}}, \textcolor{blue}{\textbf{blue}} and \textbf{bold}.}
\label{tab:MARS&Duke}
\resizebox{0.92\textwidth}{!}{
\begin{tabular}{l|c|c c c c|c c c c}
\hline
\multirow{2}{*}{\textbf{Methods}} & \multirow{2}{*}{\textbf{Backbone}} 
& \multicolumn{4}{c|}{\textbf{MARS-Attribute Dataset}} 
& \multicolumn{4}{c}{\textbf{DukeMTMC-VID-Attribute Dataset}} \\
\cline{3-10}
& & \textbf{Acc} & \textbf{Prec} & \textbf{Recall} & \textbf{F1} 
  & \textbf{Acc} & \textbf{Prec} & \textbf{Recall} & \textbf{F1} \\
\hline
VTB~\cite{cheng2022simple}             & ViT-B/16 & \textbf{72.73} & \textbf{82.79} & \textcolor{blue}{\textbf{83.52}} & \textbf{82.89} & \textbf{72.50} & \textbf{83.60} &\textcolor{blue}{\textbf{82.24}} & \textbf{82.38} \\
Zhou et al.~\cite{zhou2023solution}     & ConvNext & 69.43 & 81.75 & 80.64 & 81.19 & 65.51 & 78.42 & 77.40 & 77.40  \\
VTB-PLIP~\cite{zuo2024plip}        & ResNet50 & 54.93 & 70.02 & 69.23 & 69.14 & 43.95 & 60.80 & 58.52 & 59.10  \\
Rethink-PLIP~\cite{zuo2024plip}    & ResNet50 & 47.85 & 62.15 & 64.80 & 63.45 & 40.25 & 55.41 & 56.13 & 55.77  \\
SSPNet~\cite{zhou2024pedestrian}          & ResNet50 & 65.68 & 77.09 & 79.74 & 78.39 & 67.53 & 78.54 & 79.78 & 79.15  \\
MambaPAR~\cite{wang2024state}        & Vim & 69.07 & 81.83 & 79.39 & 80.24 & 61.55 & 75.54 & 74.84 & 74.24  \\
OTN-RWKV~\cite{wang2025rgb} & RWKV-B & \textcolor{blue}{\textbf{73.21}} & \textcolor{red}{\textbf{85.63}} & \textbf{81.53} & \textcolor{blue}{\textbf{83.22}} & \textcolor{blue}{\textbf{73.15}} & \textcolor{red}{\textbf{84.45}} & \textbf{82.16} & \textcolor{blue}{\textbf{82.78}}  \\
\hline
Baseline & RWKV-B & 72.58 & 82.81 & 83.42 & 82.82 & 72.57 & 82.36 & 83.38 & 82.38  \\
EvRainDrop (Ours) & RWKV-B, HG & 
\textcolor{red}{\textbf{73.73}} & \textcolor{blue}{\textbf{83.82}} & \textcolor{red}{\textbf{83.87}} & \textcolor{red}{\textbf{83.57}} & 
\textcolor{red}{\textbf{74.01}} & \textcolor{blue}{\textbf{84.06}} & \textcolor{red}{\textbf{83.40}} & \textcolor{red}{\textbf{83.32}}  \\
\hline
\end{tabular}
}
\end{table*}

\begin{table}
\centering
\caption{Results on the PokerEvent~\cite{wang2025sstformer} dataset. The best three results are highlighted in \textcolor{red}{\textbf{red}}, \textcolor{blue}{\textbf{blue}} and \textbf{bold}.}
\label{tab:Poker}
\resizebox{0.49\textwidth}{!}{
\begin{tabular}{c|l|c|c|c}
\hline
\textbf{\#Index} & \textbf{Methods} & \textbf{Source} & \textbf{Backbone} & \textbf{Top-1}  \\
\hline
01& C3D~\cite{ji20123d}             & ICCV-2015 & 3D CNN & 51.76  \\
02& TSM~\cite{lin2019tsm}        & ICCV-2019 & ResNet50 & 55.43  \\
03& ACTION-Net~\cite{wang2021action}             & CVPR-2021 & ResNet50 & 54.29  \\
04& TAM~\cite{liu2021tam}             & ICCV-2021 & ResNet50 & 53.65  \\
05& V-SwinTrans~\cite{liu2022video}        & CVPR-2022 & Swin-Former & 54.17  \\
06& TimeSformer~\cite{bertasius2021space}             & ICML-2021 & ViT & \textbf{55.69}  \\
07& X3D~\cite{feichtenhofer2020x3d}             & CVPR-2020 & ResNet & 51.75  \\
08& MVIT~\cite{li2022mvitv2}  & CVPR-2022 & ViT & 55.02  \\
09& SCNN-MST~\cite{wang2025sstformer} & TCDS-2025 & SCNN-Former & 53.19  \\
10& Spikingformer-MST~\cite{wang2025sstformer} & TCDS-2025 & SNN-Former & 54.74  \\
11& EFV++~\cite{chen2024retain} & TMM-2025 & Former-GNN & 55.40  \\
12& TSCFormer~\cite{wang2023unleashing} & MIR-2025 & TSCFormer & \textcolor{red}{\textbf{57.70}}  \\
\hline
13& Baseline & - & ResNet & 55.37  \\
14& EvRainDrop (Ours) & - & ResNet, HG & \textcolor{blue}{\textbf{57.62}}  \\
\hline
\end{tabular}
}
\end{table}

\section{Experiments} 
\label{sec:experiments}

\subsection{Datasets and Evaluation Metrics}  
In this study, we evaluate our proposed model on four benchmark datasets: two for single-label classification, i.e.,  \textbf{PokerEvent}~\cite{wang2025sstformer} and \textbf{HARDVS}~\cite{wang2024hardvs}, and two for multi-label classification, \textbf{MARS-Attribute}~\cite{zheng2016mars} and \textbf{DukeMTMC-VID-Attribute}~\cite{ristani2016performance}.
To evaluate the performance of our model and other state-of-the-art (SOTA) HAR and PAR models, we use different metrics for various datasets. For HAR, \textbf{top-1 accuracy} is used for the PokerEvent datasets, whereas both top-1 and top-5 accuracies are used for the HARDVS dataset. In the case of PAR, the MARS-Attribute and DukeMTMC-VID-Attribute datasets are evaluated using \textbf{Accuracy (Acc)}, \textbf{Precision (Prec)}, \textbf{Recall}, and \textbf{F1-score (F1)}. 

\begin{table}
\centering
\caption{Results on the HARDVS~\cite{wang2024hardvs} dataset. The best three results are highlighted in \textcolor{red}{\textbf{red}}, \textcolor{blue}{\textbf{blue}} and \textbf{bold}.}
\label{tab:HARDVS}
\resizebox{0.50\textwidth}{!}{
\begin{tabular}{c|l|c|c|c|c}
\hline
\textbf{\#Index} & \textbf{Methods} & \textbf{Source} & \textbf{Backbone} & \textbf{Top-1} & 
\textbf{Top-5}  \\
\hline
01& C3D~\cite{ji20123d} & ICCV-2015 & 3D CNN & 50.88 & 56.51 \\
02& R2Plus1D~\cite{tran2018closer} & CVPR-2018 & ResNet34 & 49.06 & 56.43 \\
03& SlowFast~\cite{feichtenhofer2019slowfast} & ICCV-2019 & ResNet50 & 46.54 & 54.76 \\
04& TSM~\cite{lin2019tsm} & ICCV-2019 & ResNet50 & 52.58 & \textbf{62.12} \\
05& TimeSformer~\cite{bertasius2021space} & ICML-2021 & ViT & 51.57 & 58.48 \\
06& X3D~\cite{feichtenhofer2020x3d} & CVPR-2020 & ResNet & 47.38 & 51.42 \\
07& ACTION-Net~\cite{wang2021action}             & CVPR-2021 & ResNet50 & 46.85 & 56.19 \\
08& V-SwinTrans~\cite{liu2022video} & CVPR-2022 & Swin-Former & 51.91 & 59.11 \\
09& ESTF~\cite{wang2024hardvs} & AAAI-2024 & ResNet18 & 49.93 & 55.77 \\
10& SSTFormer~\cite{wang2025sstformer} & TCDS-2025 & SNN-Former & \textcolor{blue}{\textbf{52.97}} & 60.17 \\
11& TSCFormer~\cite{wang2023unleashing} & MIR-2025 & TSCFormer & \textcolor{red}{\textbf{53.04}} & \textcolor{blue}{\textbf{62.67}} \\
\hline
12& EvRainDrop (Ours) & - & ResNet, HG & \textbf{52.60} & \textcolor{red}{\textbf{62.86}} \\
\hline
\end{tabular}
}
\end{table}

\noindent $\bullet$ \textbf{PokerEvent Dataset}~\cite{wang2025sstformer} is a sizeable RGB-Event pattern recognition benchmark dataset that focuses on character patterns in poker cards, recorded using a DAVIS 346 event camera with a spatial resolution of 346×260, containing 114 classes and 27,102 frame-event pairs, which are split into 16,216 training samples, 2,687 validation samples and 8,199 testing samples.

\noindent $\bullet$ \textbf{HARDVS Dataset}~\cite{wang2024hardvs} is a massive benchmark dataset for event-based human activity recognition, containing 300 classes of daily human activities, more than 107,646 event sequences each lasting about 5-10 seconds, considering multiple challenging factors such as multi-views, multi-illumination, multi-motion, dynamic background, occlusion and different capture distances, and split into 64,526 training samples, 10,734 validation samples and 32,386 testing samples.

\noindent $\bullet$ \textbf{MARS-Attribute Dataset}~\cite{zheng2016mars} is designed for pedestrian attribute recognition, with attributes such as actions, orientations, clothing colors, and gender decomposed into 43 binary categories. It includes 8,298 training tracklets with about 60 frames each and 625 unique identities, and 8,062 testing tracklets with around 60 frames each and 626 identities.

\begin{figure*}
\centering
\includegraphics[width=0.94\linewidth]{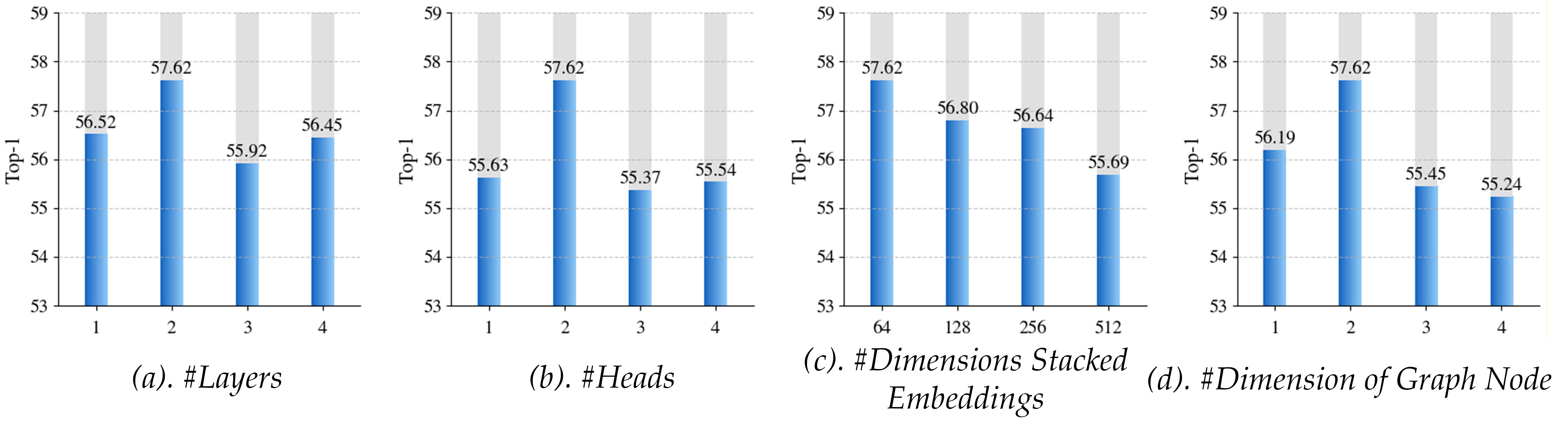}
\caption{Hyperparameter analysis of our proposed framework on the PokerEvent dataset.} 
\label{fig:hyperparameterIMG}
\end{figure*}

\begin{figure*}
\centering
\includegraphics[width=1\linewidth]{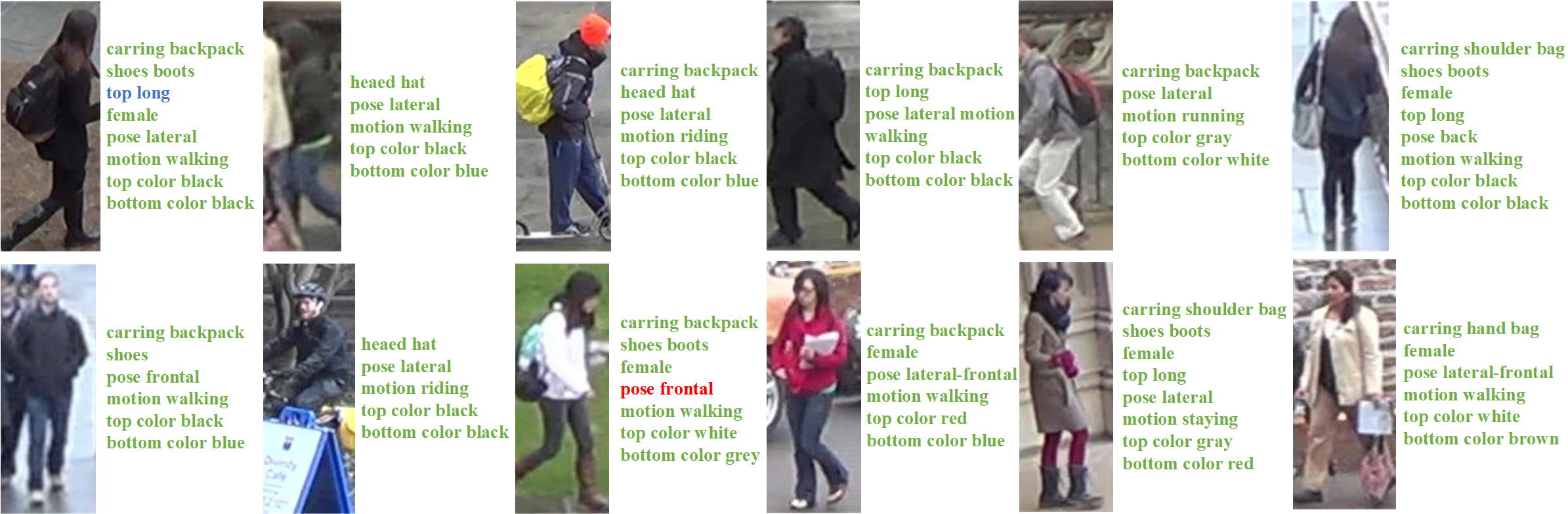}
\caption{Visualization of pedestrian attributes predicted by our proposed method on DukeMTMC-VID-Attribute Dataset. The \textcolor{red}{red} attributes indicate incorrect predictions, \textcolor{blue}{blue} attributes indicate missing predictions, and green attributes represent the ground truth.} 
\label{fig:Vis_PAR}
\end{figure*}

\begin{table}
\centering
\caption{Ablation study on the PokerEvent dataset. Symbols ``$\times$'' and ``$\checkmark$'' indicate the exclusion and inclusion of specific modules, respectively.}
\label{tab:ablation_component}
\resizebox{0.34\textwidth}{!}{
\begin{tabular}{l|ccc|c}
\hline
Method & ST1 & HGC & ST2 & Top-1 \\
\hline
Baseline & $\times$ & $\times$ & $\times$ & 55.37  \\
(a) & $\checkmark$ & $\times$ & $\times$ & 56.08  \\
(b) & $\checkmark$ & $\checkmark$ & $\times$ & 56.73 \\
\hline
(c) & $\checkmark$ & $\checkmark$ & $\checkmark$ & \textbf{57.62} \\
\hline
\end{tabular}
}
\end{table}\textbf{}

\noindent $\bullet$ \textbf{DukeMTMC-VID-Attribute Dataset}~\cite{ristani2016performance} builds on the DukeMTMC-VID dataset, tailored for pedestrian attribute recognition to enhance pedestrian re-identification performance via detailed attribute annotations. It comprises 2,032 unique pedestrian identities and 16,522 video sequences captured in diverse, challenging real-world scenarios. Its rich labeling stands as a core feature, with multi-label attributes converted into 36 binary versions for training and testing processes. The training subset has 702 identities and 16,522 images, while the testing subset includes 17,661 images of the same 702 identities. Wang et al.~\cite{wang2025rgb} further extended the MARS-Attribute and DukeMTMC-VID-Attribute datasets by incorporating simulated event data that aligned with the original RGB samples, aiming to enhance the completeness and fairness of the PAR research.

\subsection{Implementation Details}  
For both HAR and PAR we adopt consistent implementation settings where applicable. Both tasks use SGD as the optimizer with initial learning rates set to 0.0005 and 0.003, respectively. Training is conducted for 35 epochs for HAR and 100 epochs for PAR with batch sizes of 5 and 6, respectively. Input images for both tasks are resized to 224×224 and 256×128 with random horizontal flipping applied during training. Our model is implemented in PyTorch~\cite{paszke2019pytorch} and evaluated on a server with NVIDIA RTX 3090 GPUs.

\subsection{Comparison on Public Benchmarks} 

\noindent $\bullet$ \textbf{Results on PokerEvent Dataset~\cite{wang2025sstformer}.} As presented in Table~\ref{tab:Poker}, our model is compared against 12 SOTA algorithms on the PokerEvent dataset. It is evident that our model reaches a top-1 score of 57.62\%, outperforming the most recent recognition approaches. These comparative results fully validate the effectiveness of the proposed modules in our model for HAR tasks.

\noindent $\bullet$ \textbf{Results on HARDVS Dataset~\cite{wang2024hardvs}.} As shown in Table~\ref{tab:HARDVS}, our method demonstrates highly competitive performance and exceptional robustness, which is specifically designed to test models under adverse conditions with noisy event data.
Specifically, EvRainDrop achieves a Top-1 accuracy of 52.60\%, placing it on par with the SOTA methods TSCFormer~\cite{wang2023unleashing} and SSTFormer~\cite{wang2025sstformer}.
Notably, our model's strength becomes more apparent in the Top-5 accuracy, where it secures the best result with 62.86\%, outperforming all other prior works.

\noindent $\bullet$ \textbf{Results on MARS-Attribute~\cite{zheng2016mars} and DukeMTMC-VID-Attribute Dataset~\cite{ristani2016performance}.} As illustrated in Table~\ref{tab:MARS&Duke}, EvRainDrop consistently achieves new SOTA in overall acc and F1, validating its effectiveness in complex, real-world scenarios. While OTN-RWKV~\cite{wang2025rgb} attains a higher Prec, our method achieves a significantly better Recall. This suggests that our method is more effective at identifying the complete set of attributes for a person, a crucial capability for comprehensive understanding. The superior balance between Prec and Recall culminates in the highest F1.

\begin{table}
\centering
\caption{Comparison of Different Hypergraph Encoders and Aggregation Methods.}
\label{tab:HGEncoder&Aggregation}
\resizebox{0.45\textwidth}{!}{
\begin{tabular}{l|c|l|c}
\hline
HG Encoders & Top-1 & Aggregation & Top-1 \\
\hline
UniGIN~\cite{huang2021unignn} & 56.41 & Concat Fusion  & 56.57 \\
UniGCN~\cite{huang2021unignn}  & 55.65 & Weighted Fusion  & 55.71 \\
UniGCN2~\cite{huang2021unignn}  & 56.35 & Hierarchical Fusion  & 56.02 \\
UniGAT  & 57.62 & Concat all nodes & 57.62 \\
\hline
\end{tabular}
}
\end{table}\textbf{}

\begin{table}
\centering
\caption{Ablation study on $k$ for Selecting top-$k$ Static Nodes in Hyperedge Formation.}
\label{tab:topk}
\small
\resizebox{0.40\textwidth}{!}{
\begin{tabular}{c|cccccccc}
\hline
$k$ & 4 & 6 & 8 & 10 & 12 \\
\hline
Top-1 & 56.19 & 57.62 & 55.76 & 56.37 & 56.49 \\
\hline
\end{tabular}
}
\end{table}

\subsection{Ablation Study} 
In order to validate the components of our approach, a comprehensive ablation study was carried out on the proposed EvRainDrop using the PokerEvent dataset.

\noindent $\bullet$ \textbf{Component Analysis.} 
As shown in Table~\ref{tab:ablation_component}, we evaluate three key components’ contributions to the model’s accuracy on the PokerEvent dataset: Hypergraph Propagation’s two stages (ST1, ST2) and Hypergraph Construction (HGC).
Incorporating ST1 improves Top-1 accuracy from 55.37\% to 56.08\%, showing that it captures dynamic node relationships and enhances self-completion. Adding HGC further boosts accuracy, confirming the benefit of hypergraph construction. Finally, ST2 raises Top-1 from 56.73\% to 57.62\%, indicating that cross-modality hypergraph enhancement improves information transfer.

\begin{figure}
\centering
\includegraphics[width=1\linewidth]{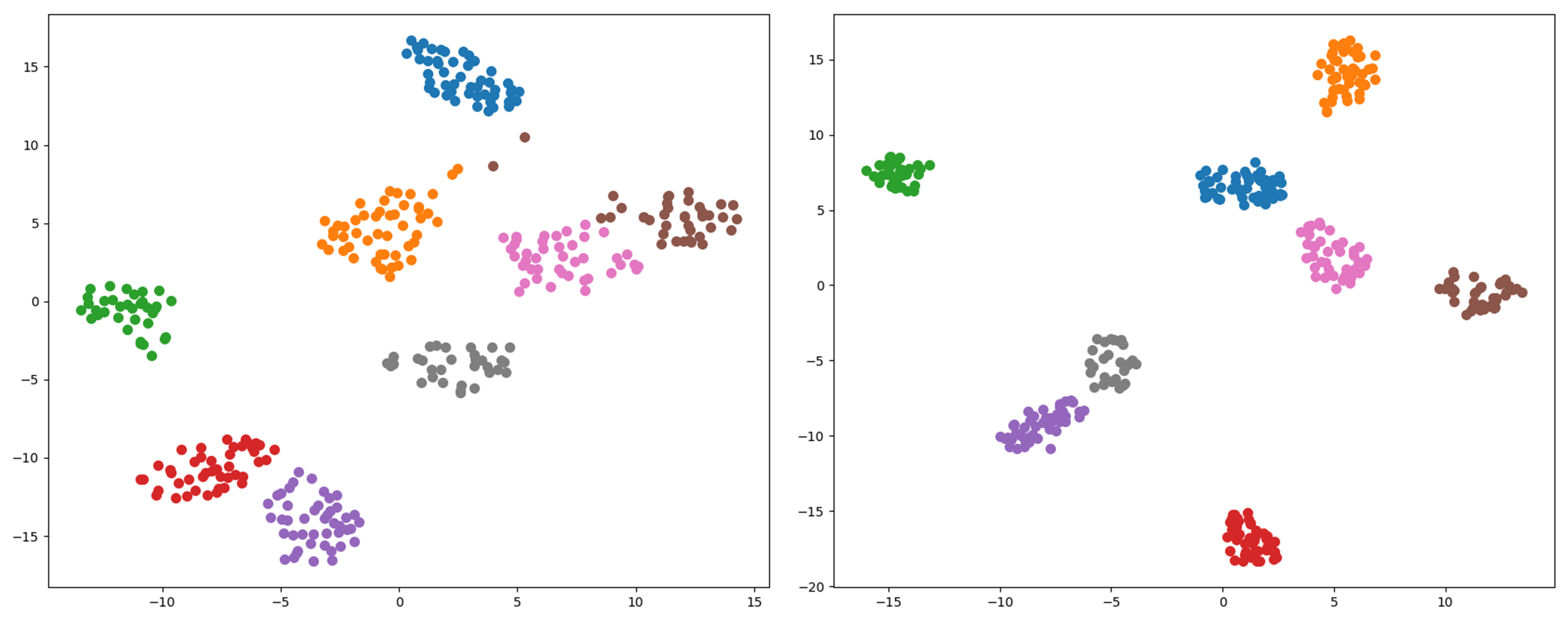}
\caption{The feature distribution of Baseline (left sub-figure) and our newly proposed model (right sub-figure) on the PokerEvent dataset using T-SNE.} 
\label{fig:tsne}
\end{figure}

\begin{figure}
\centering
\includegraphics[width=1.0\linewidth]{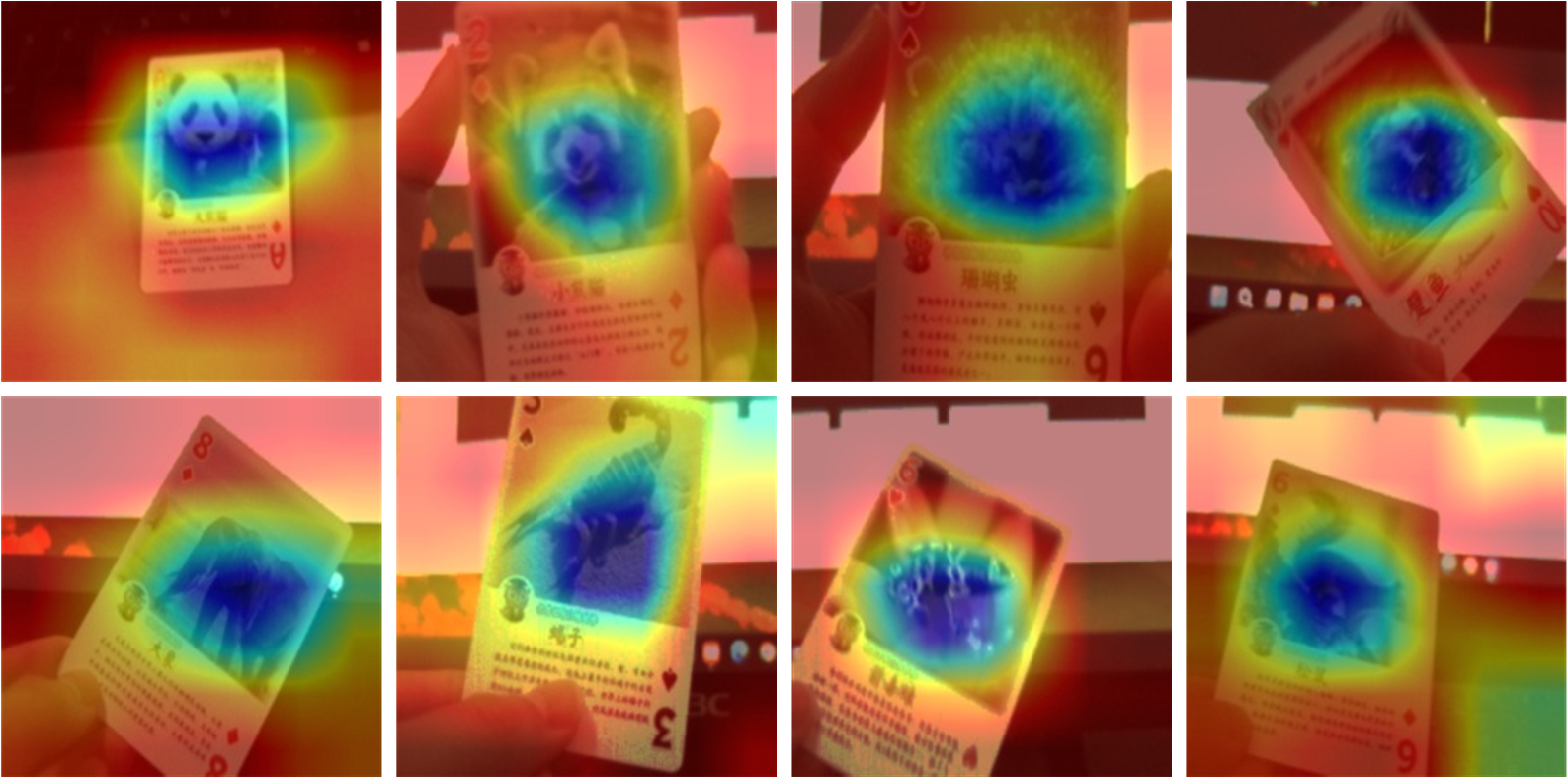}
\caption{Visualization of feature maps obtained using our proposed EvRainDrop on the PokerEvent datasets.} 
\label{fig:vis_map}
\end{figure}

\noindent $\bullet$ \textbf{Effect of Different Hypergraph Encoders.} As shown in Table~\ref{tab:HGEncoder&Aggregation}, we perform ablation studies with four encoders derived from the UniGNN~\cite{huang2021unignn}, and design our method in a UniGAT like style. UniGIN generalizes Graph Isomorphism Networks to hypergraphs by balancing self-features and hyperedge information. UniGCN adapts Graph Convolutional Networks by aggregating node embeddings with degree-based hyperedge weights. UniGCNII mitigates over-smoothing through residual connections and identity mapping.

\noindent $\bullet$ \textbf{Impact of Hypergraph Node Aggregation Methods.} As shown in Table~\ref{tab:HGEncoder&Aggregation}, Concat all nodes builds a single hypergraph from all nodes, while other methods first build dynamic and static node hypergraphs separately before fusion. Concat Fusion directly concatenates the hyperedge indices of the two hypergraphs. Weighted Fusion uses learnable weights to adaptively balance their contributions. Hierarchical Fusion sequentially propagates information through the two hypergraphs.

\noindent $\bullet$ \textbf{Analysis of the Parameter $k$ in Hyperedge Construction.} As shown in Table~\ref{tab:topk}, the parameter $k$ controls the number of static nodes combined with a dynamic node to form a hyperedge, thus affecting the granularity of multimodal information aggregation. We vary $k$ from 4 to 12 and report the Top-1 accuracy on PokerEvent.

\subsection{Parameter Analysis} 
As shown in Fig.~\ref{fig:hyperparameterIMG}, we use a Transformer for temporal information fusion, where nhead, nlayer, and nhid denote the number of parallel attention heads in multi-head attention, the number of stacked Transformer encoder layers, and the hidden dimension of the feedforward network in each encoder layer, respectively. $d_{ob}$ denotes the feature split dimension for each dynamic node in EvRainDrop, matching the hypergraph propagation layer’s input dimension requirement and enabling feature expansion for dynamic temporal nodes.

\begin{figure}
\centering
\includegraphics[width=1.0\linewidth]{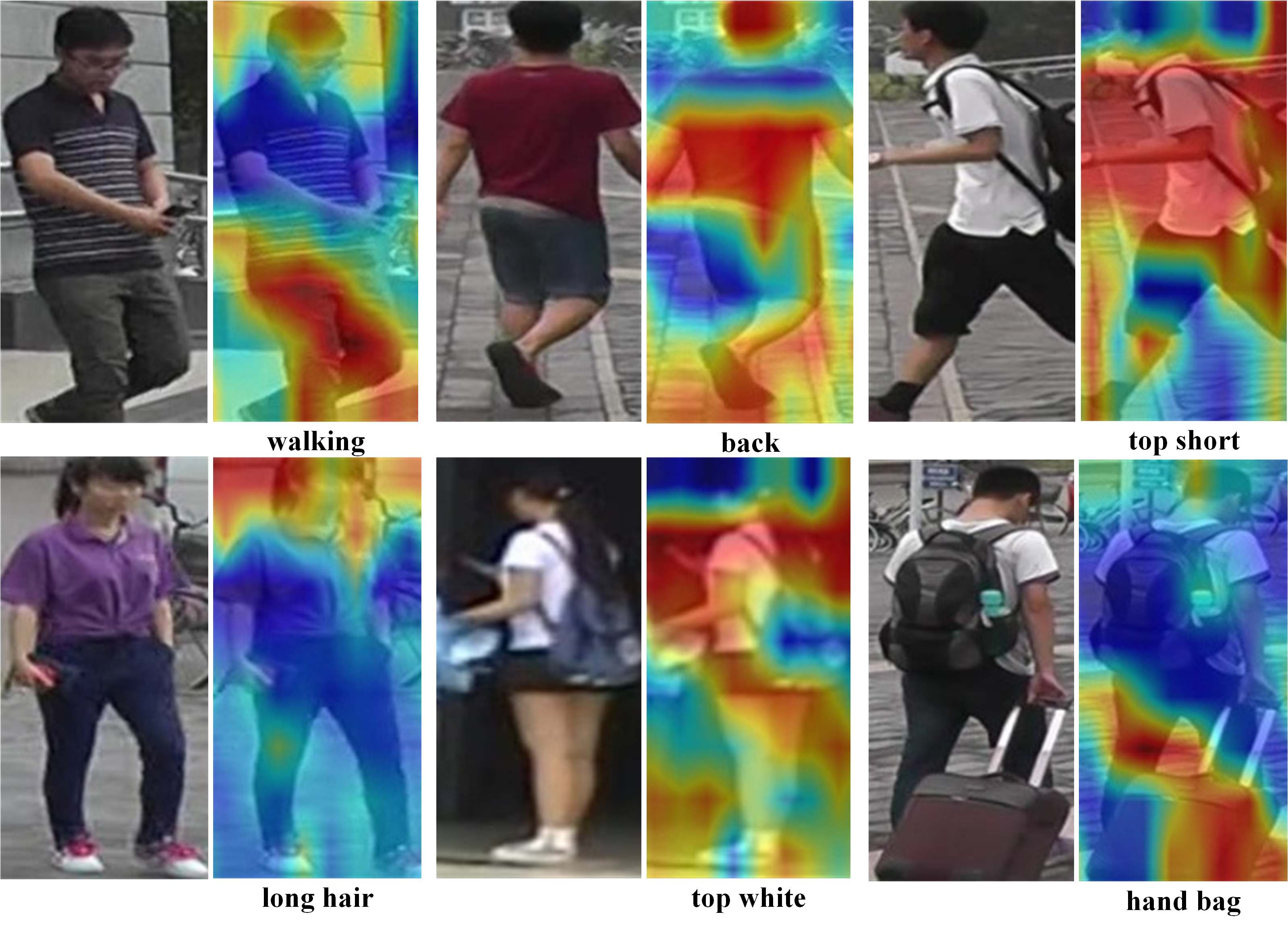}
\caption{Visualization of heat maps given the corresponding pedestrian attribute on the MARS-Attribute datasets.} 
\label{fig:vis_map_mars}
\end{figure}

\subsection{Visualization} 

\noindent $\bullet$ \textbf{Attributes Predicted.} As shown in Fig.~\ref{fig:Vis_PAR}, this study presents PAR results on the DukeMTMC-VID-Attribute dataset. Both quantitative and qualitative evaluations show the model’s stability and accuracy in identifying key attributes such as gender, pose, motion, and clothing.

\textbf{Distribution and Feature Maps.} As shown in Fig.~\ref{fig:tsne}, we visualize feature embeddings of the Baseline and our proposed method on PokerEvent. Each color denotes a distinct test class. The Baseline shows scattered clusters with low intra-class compactness and small inter-class margins, indicating difficulty in grouping similar samples and frequent class confusion. In contrast, EvRainDrop forms a well-structured feature space with compact clusters and clear separations, showing strong intra-class cohesion and inter-class discrimination.

As shown in Fig.~\ref{fig:vis_map}, we give the feature maps of our EvRainDrop. It can be observed that our model achieves highly focused attention on the Poker samples. As shown in Fig.~\ref{fig:vis_map_mars}, on the MARS-Attribute dataset, the heatmap indicates that during inference, EvRainDrop can effectively attend to semantic regions closely associated with various attributes.
    
\subsection{Limitation Analysis}  
Despite the promising performance of EvRainDrop on HAR and PAR, it has limitations. For downstream tasks, the current multimodal fusion mainly relies on simple concatenation and self-attention mechanisms. These effective operations may not fully exploit the intrinsic complementary relationships between event streams and RGB frames. More in-depth fusion designs remain underexplored and could further enhance the model’s ability to integrate spatiotemporal dynamics and dense spatial semantics.

\section{Conclusion}  
In this paper, we propose EvRainDrop, a hypergraph-guided framework for spatiotemporal completion and aggregation of event streams and RGB frames. Inspired by raindrops’ asynchronous nature, it models event and RGB tokens as hypergraph nodes, using hypergraph propagation to alleviate event data’s spatial sparsity while preserving rich temporal dynamics. Extended to multimodal scenarios, it enables robust RGB-Event information completion and fusion. Comprehensive experiments on four benchmark datasets for both single-label and multi-label classification tasks validate that EvRainDrop achieves state-of-the-art performance, demonstrating its effectiveness in event-based perception tasks.


{
    \small
    \bibliographystyle{ieeenat_fullname}
    \bibliography{main}

\begin{thebibliography}{51}
\providecommand{\natexlab}[1]{#1}
\providecommand{\url}[1]{\texttt{#1}}
\expandafter\ifx\csname urlstyle\endcsname\relax
  \providecommand{\doi}[1]{doi: #1}\else
  \providecommand{\doi}{doi: \begingroup \urlstyle{rm}\Url}\fi

\bibitem[Bertasius et~al.(2021)Bertasius, Wang, and
  Torresani]{bertasius2021space}
Gedas Bertasius, Heng Wang, and Lorenzo Torresani.
\newblock Is space-time attention all you need for video understanding?
\newblock In \emph{Icml}, page~4, 2021.

\bibitem[Chen et~al.(2023)Chen, Chung, Tan, and Chen]{chen2023dense}
Haodong Chen, Vera Chung, Li Tan, and Xiaoming Chen.
\newblock Dense voxel 3d reconstruction using a monocular event camera.
\newblock \emph{arXiv preprint arXiv:2309.00385}, 2023.

\bibitem[Chen et~al.(2020)Chen, Lei, Song, Ying, Chen, and
  Wu]{chen2020hierarchical}
Jintai Chen, Biwen Lei, Qingyu Song, Haochao Ying, Danny~Z. Chen, and Jian Wu.
\newblock A hierarchical graph network for 3d object detection on point clouds.
\newblock In \emph{2020 IEEE/CVF Conference on Computer Vision and Pattern
  Recognition (CVPR)}, pages 389--398, 2020.

\bibitem[Chen et~al.(2024)Chen, Li, Wang, Shao, Zhang, Wang, Tian, and
  Tang]{chen2024retain}
Lan Chen, Dong Li, Xiao Wang, Pengpeng Shao, Wei Zhang, Yaowei Wang, Yonghong
  Tian, and Jin Tang.
\newblock Retain, blend, and exchange: A quality-aware spatial-stereo fusion
  approach for event stream recognition.
\newblock \emph{arXiv preprint arXiv:2406.18845}, 2024.

\bibitem[Chen et~al.(2019)Chen, Rohrbach, Yan, Shuicheng, Feng, and
  Kalantidis]{chen2018graphbased}
Yunpeng Chen, Marcus Rohrbach, Zhicheng Yan, Yan Shuicheng, Jiashi Feng, and
  Yannis Kalantidis.
\newblock Graph-based global reasoning networks.
\newblock In \emph{Proceedings of the IEEE/CVF conference on computer vision
  and pattern recognition}, pages 433--442, 2019.

\bibitem[Cheng et~al.(2022)Cheng, Jia, Wang, and Zhang]{cheng2022simple}
Xinhua Cheng, Mengxi Jia, Qian Wang, and Jian Zhang.
\newblock A simple visual-textual baseline for pedestrian attribute
  recognition.
\newblock \emph{IEEE Transactions on Circuits and Systems for Video
  Technology}, 32\penalty0 (10):\penalty0 6994--7004, 2022.

\bibitem[Feichtenhofer(2020)]{feichtenhofer2020x3d}
Christoph Feichtenhofer.
\newblock X3d: Expanding architectures for efficient video recognition.
\newblock In \emph{Proceedings of the IEEE/CVF conference on computer vision
  and pattern recognition}, pages 203--213, 2020.

\bibitem[Feichtenhofer et~al.(2019)Feichtenhofer, Fan, Malik, and
  He]{feichtenhofer2019slowfast}
Christoph Feichtenhofer, Haoqi Fan, Jitendra Malik, and Kaiming He.
\newblock Slowfast networks for video recognition.
\newblock In \emph{Proceedings of the IEEE/CVF international conference on
  computer vision}, pages 6202--6211, 2019.

\bibitem[Huang and Yang(2021)]{huang2021unignn}
Jing Huang and Jie Yang.
\newblock Unignn: a unified framework for graph and hypergraph neural networks.
\newblock \emph{arXiv preprint arXiv:2105.00956}, 2021.

\bibitem[Ji et~al.(2012)Ji, Xu, Yang, and Yu]{ji20123d}
Shuiwang Ji, Wei Xu, Ming Yang, and Kai Yu.
\newblock 3d convolutional neural networks for human action recognition.
\newblock \emph{IEEE transactions on pattern analysis and machine
  intelligence}, 35\penalty0 (1):\penalty0 221--231, 2012.

\bibitem[Kai et~al.(2024)Kai, Lu, Zhang, and Sun]{kai2024evtexture}
Dachun Kai, Jiayao Lu, Yueyi Zhang, and Xiaoyan Sun.
\newblock Evtexture: event-driven texture enhancement for video
  super-resolution.
\newblock \emph{arXiv preprint arXiv:2406.13457}, 2024.

\bibitem[Kai et~al.(2025)Kai, Zhang, Wang, Xiao, Xiong, and
  Sun]{kai2025eventenhanced}
Dachun Kai, Yueyi Zhang, Jin Wang, Zeyu Xiao, Zhiwei Xiong, and Xiaoyan Sun.
\newblock Event-enhanced blurry video super-resolution.
\newblock In \emph{Proceedings of the AAAI Conference on Artificial
  Intelligence}, pages 4175--4183, 2025.

\bibitem[Kar et~al.(2025)Kar, Raj, and Su]{kar2025eventcameraguidedvisual}
Aupendu Kar, Vishnu Raj, and Guan-Ming Su.
\newblock Event camera guided visual media restoration \& 3d reconstruction: A
  survey.
\newblock \emph{arXiv preprint arXiv:2509.09971}, 2025.

\bibitem[Li et~al.(2022)Li, Wu, Fan, Mangalam, Xiong, Malik, and
  Feichtenhofer]{li2022mvitv2}
Yanghao Li, Chao-Yuan Wu, Haoqi Fan, Karttikeya Mangalam, Bo Xiong, Jitendra
  Malik, and Christoph Feichtenhofer.
\newblock Mvitv2: Improved multiscale vision transformers for classification
  and detection.
\newblock In \emph{Proceedings of the IEEE/CVF conference on computer vision
  and pattern recognition}, pages 4804--4814, 2022.

\bibitem[Lin et~al.(2019)Lin, Gan, and Han]{lin2019tsm}
Ji Lin, Chuang Gan, and Song Han.
\newblock Tsm: Temporal shift module for efficient video understanding.
\newblock In \emph{Proceedings of the IEEE/CVF international conference on
  computer vision}, pages 7083--7093, 2019.

\bibitem[Lin et~al.(2023)Lin, Qiu, Shen, Zang, Liu, Bian, M{\"u}ller, Wang,
  et~al.]{lin2023e2pnet}
Xiuhong Lin, Changjie Qiu, Siqi Shen, Yu Zang, Weiquan Liu, Xuesheng Bian,
  Matthias M{\"u}ller, Cheng Wang, et~al.
\newblock E2pnet: event to point cloud registration with spatio-temporal
  representation learning.
\newblock \emph{Advances in Neural Information Processing Systems},
  36:\penalty0 18076--18089, 2023.

\bibitem[Liu et~al.(2021{\natexlab{a}})Liu, Xing, Tang, Ma, and
  Pan]{liu2021event}
Qianhui Liu, Dong Xing, Huajin Tang, De Ma, and Gang Pan.
\newblock Event-based action recognition using motion information and spiking
  neural networks.
\newblock In \emph{IJCAI}, pages 1743--1749, 2021{\natexlab{a}}.

\bibitem[Liu et~al.(2021{\natexlab{b}})Liu, Wang, Wu, Qian, and Lu]{liu2021tam}
Zhaoyang Liu, Limin Wang, Wayne Wu, Chen Qian, and Tong Lu.
\newblock Tam: Temporal adaptive module for video recognition.
\newblock In \emph{Proceedings of the IEEE/CVF international conference on
  computer vision}, pages 13708--13718, 2021{\natexlab{b}}.

\bibitem[Liu et~al.(2022)Liu, Ning, Cao, Wei, Zhang, Lin, and Hu]{liu2022video}
Ze Liu, Jia Ning, Yue Cao, Yixuan Wei, Zheng Zhang, Stephen Lin, and Han Hu.
\newblock Video swin transformer.
\newblock In \emph{Proceedings of the IEEE/CVF conference on computer vision
  and pattern recognition}, pages 3202--3211, 2022.

\bibitem[Luo et~al.(2020)Luo, Li, Yang, Jiao, Cheng, and Lyu]{luo2020cascade}
Ao Luo, Xin Li, Fan Yang, Zhicheng Jiao, Hong Cheng, and Siwei Lyu.
\newblock Cascade graph neural networks for rgb-d salient object detection.
\newblock In \emph{European conference on computer vision}, pages 346--364.
  Springer, 2020.

\bibitem[Mei et~al.(2024)Mei, Bi, Wen, Kong, and Wu]{mei2024MaHGNN}
Zhangyu Mei, Xiao Bi, Yating Wen, Xianchun Kong, and Hao Wu.
\newblock Ma-hgnn+: A hypergraph neural network for depth analysis of
  multimodal data.
\newblock In \emph{2024 IEEE International Conference on Medical Artificial
  Intelligence (MedAI)}, pages 501--508, 2024.

\bibitem[Paszke et~al.(2019)Paszke, Gross, Massa, Lerer, Bradbury, Chanan,
  Killeen, Lin, Gimelshein, Antiga, et~al.]{paszke2019pytorch}
Adam Paszke, Sam Gross, Francisco Massa, Adam Lerer, James Bradbury, Gregory
  Chanan, Trevor Killeen, Zeming Lin, Natalia Gimelshein, Luca Antiga, et~al.
\newblock Pytorch: An imperative style, high-performance deep learning library.
\newblock \emph{Advances in neural information processing systems}, 32, 2019.

\bibitem[Rebecq et~al.(2021)Rebecq, Ranftl, Koltun, and
  Scaramuzza]{rebe2021vieddo}
Henri Rebecq, René Ranftl, Vladlen Koltun, and Davide Scaramuzza.
\newblock High speed and high dynamic range video with an event camera.
\newblock \emph{IEEE Transactions on Pattern Analysis and Machine
  Intelligence}, 43\penalty0 (6):\penalty0 1964--1980, 2021.

\bibitem[Ren et~al.(2024{\natexlab{a}})Ren, Zhou, Zhu, Fu, Huang, Lin, Fang,
  Ma, Yu, and Cheng]{ren2024rethinking}
Hongwei Ren, Yue Zhou, Jiadong Zhu, Haotian Fu, Yulong Huang, Xiaopeng Lin,
  Yuetong Fang, Fei Ma, Hao Yu, and Bojun Cheng.
\newblock Rethinking efficient and effective point-based networks for event
  camera classification and regression: Eventmamba.
\newblock \emph{arXiv preprint arXiv:2405.06116}, 2024{\natexlab{a}}.

\bibitem[Ren et~al.(2024{\natexlab{b}})Ren, Zhu, Zhou, Fu, Huang, and
  Cheng]{ren2024simple}
Hongwei Ren, Jiadong Zhu, Yue Zhou, Haotian Fu, Yulong Huang, and Bojun Cheng.
\newblock A simple and effective point-based network for event camera 6-dofs
  pose relocalization.
\newblock In \emph{Proceedings of the IEEE/CVF Conference on Computer Vision
  and Pattern Recognition}, pages 18112--18121, 2024{\natexlab{b}}.

\bibitem[Ristani et~al.(2016)Ristani, Solera, Zou, Cucchiara, and
  Tomasi]{ristani2016performance}
Ergys Ristani, Francesco Solera, Roger Zou, Rita Cucchiara, and Carlo Tomasi.
\newblock Performance measures and a data set for multi-target, multi-camera
  tracking.
\newblock In \emph{European conference on computer vision}, pages 17--35.
  Springer, 2016.

\bibitem[Schaefer et~al.(2022)Schaefer, Gehrig, and
  Scaramuzza]{schaefer2022aegnn}
Simon Schaefer, Daniel Gehrig, and Davide Scaramuzza.
\newblock Aegnn: Asynchronous event-based graph neural networks.
\newblock In \emph{Proceedings of the IEEE/CVF conference on computer vision
  and pattern recognition}, pages 12371--12381, 2022.

\bibitem[Tang et~al.(2022)Tang, Wang, Huang, Jiang, Zhu, Zhang, Wang, and
  Tian]{tang2022revisiting}
Chuanming Tang, Xiao Wang, Ju Huang, Bo Jiang, Lin Zhu, Jianlin Zhang, Yaowei
  Wang, and Yonghong Tian.
\newblock Revisiting color-event based tracking: A unified network, dataset,
  and metric.
\newblock \emph{arXiv preprint arXiv:2211.11010}, 2022.

\bibitem[Tran et~al.(2018)Tran, Wang, Torresani, Ray, LeCun, and
  Paluri]{tran2018closer}
Du Tran, Heng Wang, Lorenzo Torresani, Jamie Ray, Yann LeCun, and Manohar
  Paluri.
\newblock A closer look at spatiotemporal convolutions for action recognition.
\newblock In \emph{Proceedings of the IEEE conference on Computer Vision and
  Pattern Recognition}, pages 6450--6459, 2018.

\bibitem[Wang et~al.(2023{\natexlab{a}})Wang, Li, Zhu, Zhang, Chen, Li, Wang,
  Tian, and Wu]{wang2023visevent}
Xiao Wang, Jianing Li, Lin Zhu, Zhipeng Zhang, Zhe Chen, Xin Li, Yaowei Wang,
  Yonghong Tian, and Feng Wu.
\newblock Visevent: Reliable object tracking via collaboration of frame and
  event flows.
\newblock \emph{IEEE Transactions on Cybernetics}, 54\penalty0 (3):\penalty0
  1997--2010, 2023{\natexlab{a}}.

\bibitem[Wang et~al.(2023{\natexlab{b}})Wang, Rong, Wang, Chen, Wu, Jiang,
  Tian, and Tang]{wang2023unleashing}
Xiao Wang, Yao Rong, Shiao Wang, Yuan Chen, Zhe Wu, Bo Jiang, Yonghong Tian,
  and Jin Tang.
\newblock Unleashing the power of cnn and transformer for balanced rgb-event
  video recognition.
\newblock \emph{arXiv preprint arXiv:2312.11128}, 2023{\natexlab{b}}.

\bibitem[Wang et~al.(2024{\natexlab{a}})Wang, Rong, Wang, Li, Zhu, Jiang, and
  Wang]{wang2024eventSLT}
Xiao Wang, Yao Rong, Fuling Wang, Jianing Li, Lin Zhu, Bo Jiang, and Yaowei
  Wang.
\newblock Event stream based sign language translation: A high-definition
  benchmark dataset and a new algorithm.
\newblock \emph{arXiv preprint arXiv:2408.10488}, 2024{\natexlab{a}}.

\bibitem[Wang et~al.(2024{\natexlab{b}})Wang, Wang, Ding, Li, Wu, Rong, Kong,
  Huang, Li, Yang, et~al.]{wang2024state}
Xiao Wang, Shiao Wang, Yuhe Ding, Yuehang Li, Wentao Wu, Yao Rong, Weizhe Kong,
  Ju Huang, Shihao Li, Haoxiang Yang, et~al.
\newblock State space model for new-generation network alternative to
  transformers: A survey.
\newblock \emph{arXiv preprint arXiv:2404.09516}, 2024{\natexlab{b}}.

\bibitem[Wang et~al.(2024{\natexlab{c}})Wang, Wang, Shao, Jiang, Zhu, and
  Tian]{wang2024event}
Xiao Wang, Shiao Wang, Pengpeng Shao, Bo Jiang, Lin Zhu, and Yonghong Tian.
\newblock Event stream based human action recognition: a high-definition
  benchmark dataset and algorithms.
\newblock \emph{arXiv preprint arXiv:2408.09764}, 2024{\natexlab{c}}.

\bibitem[Wang et~al.(2024{\natexlab{d}})Wang, Wu, Jiang, Bao, Zhu, Li, Wang,
  and Tian]{wang2024hardvs}
Xiao Wang, Zongzhen Wu, Bo Jiang, Zhimin Bao, Lin Zhu, Guoqi Li, Yaowei Wang,
  and Yonghong Tian.
\newblock Hardvs: Revisiting human activity recognition with dynamic vision
  sensors.
\newblock In \emph{Proceedings of the AAAI conference on artificial
  intelligence}, pages 5615--5623, 2024{\natexlab{d}}.

\bibitem[Wang et~al.(2025{\natexlab{a}})Wang, Jin, Wu, Zhang, Zhu, Jiang, and
  Tian]{wang2025object}
Xiao Wang, Yu Jin, Wentao Wu, Wei Zhang, Lin Zhu, Bo Jiang, and Yonghong Tian.
\newblock Object detection using event camera: A moe heat conduction based
  detector and a new benchmark dataset.
\newblock In \emph{Proceedings of the Computer Vision and Pattern Recognition
  Conference}, pages 29321--29330, 2025{\natexlab{a}}.

\bibitem[Wang et~al.(2025{\natexlab{b}})Wang, Li, Wang, Jiang, Wang, Tian,
  Tang, and Luo]{wang2025signLTFE}
Xiao Wang, Yuehang Li, Fuling Wang, Bo Jiang, Yaowei Wang, Yonghong Tian, Jin
  Tang, and Bin Luo.
\newblock Sign language translation using frame and event stream: Benchmark
  dataset and algorithms.
\newblock \emph{arXiv preprint arXiv:2503.06484}, 2025{\natexlab{b}}.

\bibitem[Wang et~al.(2025{\natexlab{c}})Wang, Rong, Wu, Zhu, Jiang, Tang, and
  Tian]{wang2025sstformer}
Xiao Wang, Yao Rong, Zongzhen Wu, Lin Zhu, Bo Jiang, Jin Tang, and Yonghong
  Tian.
\newblock Sstformer: Bridging spiking neural network and memory support
  transformer for frame-event based recognition.
\newblock \emph{IEEE Transactions on Cognitive and Developmental Systems},
  2025{\natexlab{c}}.

\bibitem[Wang et~al.(2025{\natexlab{d}})Wang, Wang, Wang, Chen, Jin, Song,
  Jiang, and Li]{wang2025rgb}
Xiao Wang, Haiyang Wang, Shiao Wang, Qiang Chen, Jiandong Jin, Haoyu Song, Bo
  Jiang, and Chenglong Li.
\newblock Rgb-event based pedestrian attribute recognition: A benchmark dataset
  and an asymmetric rwkv fusion framework.
\newblock \emph{arXiv preprint arXiv:2504.10018}, 2025{\natexlab{d}}.

\bibitem[Wang et~al.(2021)Wang, She, and Smolic]{wang2021action}
Zhengwei Wang, Qi She, and Aljosa Smolic.
\newblock Action-net: Multipath excitation for action recognition.
\newblock In \emph{Proceedings of the IEEE/CVF conference on computer vision
  and pattern recognition}, pages 13214--13223, 2021.

\bibitem[Xie et~al.(2024)Xie, Deng, Shao, Xu, and Li]{xie2024event}
Bochen Xie, Yongjian Deng, Zhanpeng Shao, Qingsong Xu, and Youfu Li.
\newblock Event voxel set transformer for spatiotemporal representation
  learning on event streams.
\newblock \emph{IEEE Transactions on Circuits and Systems for Video
  Technology}, 2024.

\bibitem[Xu et~al.(2019)Xu, Jiang, Liang, and Li]{Xu2019Spatial}
Hang Xu, Chenhan Jiang, Xiaodan Liang, and Zhenguo Li.
\newblock Spatial-aware graph relation network for large-scale object
  detection.
\newblock In \emph{2019 IEEE/CVF Conference on Computer Vision and Pattern
  Recognition (CVPR)}, pages 9290--9299, 2019.

\bibitem[Yang and Xu(2025)]{yang2025recent}
Mu-Rong Yang and Xin-Jian Xu.
\newblock Recent advances in hypergraph neural networks: M.-r. yang, x.-j. xu.
\newblock \emph{Journal of the Operations Research Society of China}, pages
  1--37, 2025.

\bibitem[Yao et~al.(2018)Yao, Pan, Li, and Mei]{yao2018exploring}
Ting Yao, Yingwei Pan, Yehao Li, and Tao Mei.
\newblock Exploring visual relationship for image captioning.
\newblock In \emph{Proceedings of the European conference on computer vision
  (ECCV)}, pages 684--699, 2018.

\bibitem[Zhang et~al.(2014)Zhang, Gao, Hong, Feng, Zhu, and
  Cai]{zhang2014Feature}
Luming Zhang, Yue Gao, Chaoqun Hong, Yinfu Feng, Jianke Zhu, and Deng Cai.
\newblock Feature correlation hypergraph: Exploiting high-order potentials for
  multimodal recognition.
\newblock \emph{IEEE Transactions on Cybernetics}, 44\penalty0 (8):\penalty0
  1408--1419, 2014.

\bibitem[Zhang et~al.(2021)Zhang, Zeman, Tsiligkaridis, and
  Zitnik]{zhang2021graph}
Xiang Zhang, Marko Zeman, Theodoros Tsiligkaridis, and Marinka Zitnik.
\newblock Graph-guided network for irregularly sampled multivariate time
  series.
\newblock \emph{arXiv preprint arXiv:2110.05357}, 2021.

\bibitem[Zheng et~al.(2016)Zheng, Bie, Sun, Wang, Su, Wang, and
  Tian]{zheng2016mars}
Liang Zheng, Zhi Bie, Yifan Sun, Jingdong Wang, Chi Su, Shengjin Wang, and Qi
  Tian.
\newblock Mars: A video benchmark for large-scale person re-identification.
\newblock In \emph{European conference on computer vision}, pages 868--884.
  Springer, 2016.

\bibitem[Zhou et~al.(2023)Zhou, Hu, Yu, Xu, Lu, and Cao]{zhou2023solution}
Yibo Zhou, Hai-Miao Hu, Jinzuo Yu, Zhenbo Xu, Weiqing Lu, and Yuran Cao.
\newblock A solution to co-occurrence bias: Attributes disentanglement via
  mutual information minimization for pedestrian attribute recognition.
\newblock \emph{arXiv preprint arXiv:2307.15252}, 2023.

\bibitem[Zhou et~al.(2024)Zhou, Hu, Xiang, Zhang, and Wu]{zhou2024pedestrian}
Yibo Zhou, Hai-Miao Hu, Yirong Xiang, Xiaokang Zhang, and Haotian Wu.
\newblock Pedestrian attribute recognition as label-balanced multi-label
  learning.
\newblock \emph{arXiv preprint arXiv:2405.04858}, 2024.

\bibitem[Zubic et~al.(2024)Zubic, Gehrig, and Scaramuzza]{zubic2024state}
Nikola Zubic, Mathias Gehrig, and Davide Scaramuzza.
\newblock State space models for event cameras.
\newblock In \emph{Proceedings of the IEEE/CVF Conference on Computer Vision
  and Pattern Recognition}, pages 5819--5828, 2024.

\bibitem[Zuo et~al.(2024)Zuo, Hong, Zhang, Yu, Zhou, Gao, Sang, and
  Wang]{zuo2024plip}
Jialong Zuo, Jiahao Hong, Feng Zhang, Changqian Yu, Hanyu Zhou, Changxin Gao,
  Nong Sang, and Jingdong Wang.
\newblock Plip: Language-image pre-training for person representation learning.
\newblock \emph{Advances in Neural Information Processing Systems},
  37:\penalty0 45666--45702, 2024.

\end{thebibliography}
}


\end{document}